\documentclass[letterpaper, 10 pt, conference]{ieeeconf}  

\IEEEoverridecommandlockouts                              

\usepackage{graphics} 
\usepackage{epsfig} 
\usepackage{amsmath} 
\usepackage{amssymb}  
\usepackage{multirow}
\usepackage{xcolor}
\usepackage{hyperref}
\usepackage[nocompress]{cite}

\usepackage{array}
\newcolumntype{P}[1]{>{\centering\arraybackslash}p{#1}}

\title{\LARGE \bf
Sense, Imagine, Act: Multimodal Perception Improves Model-Based Reinforcement Learning for Head-to-Head Autonomous Racing
}

\author{Elena Shrestha$^{1}$, Chetan Reddy$^{2*}$, Hanxi Wan$^{2*}$, Yulun Zhuang$^{1}$, and Ram Vasudevan$^{1}$
  \thanks{$*$ These authors contributed equally to this work}
\thanks{$^{1}$Elena Shrestha, Yulun Zhuang, and Ram Vasudevan are with the Department of Robotics, University of Michigan, Ann Arbor, MI. 
        $\langle${\tt\small eshresco,yulunz,ramv}$\rangle${\tt\small@umich.edu}}%
\thanks{$^{2}$Chetan Reddy and Hanxi Wan are with the Department of Electrical Engineering and Computer Science, University of Michigan, Ann Arbor, MI. 
  $\langle${\tt\small chereddy,wanhanxi}$\rangle${\tt\small@umich.edu}}%
}

\begin{document}

\maketitle
\thispagestyle{empty}
\pagestyle{empty}

\begin{abstract}
Model-based reinforcement learning (MBRL) techniques have recently yielded promising results for real-world autonomous racing using high-dimensional observations. MBRL agents, such as Dreamer, solve long-horizon tasks by building a world model and planning actions by latent imagination. This approach involves explicitly learning a model of the system dynamics and using it to learn the optimal policy for continuous control over multiple timesteps. As a result, MBRL agents may converge to sub-optimal policies if the world model is inaccurate. To improve state estimation for autonomous racing, this paper proposes a self-supervised sensor fusion technique that combines egocentric LiDAR and RGB camera observations collected from the F1TENTH Gym. The zero-shot performance of MBRL agents is empirically evaluated on unseen tracks and against a dynamic obstacle. This paper illustrates that multimodal perception improves robustness of the world model without requiring additional training data. The resulting multimodal Dreamer agent safely avoided collisions and won the most races compared to other tested baselines in zero-shot head-to-head autonomous racing. 
\end{abstract}


\section{INTRODUCTION}

Developing autonomous agents that learn to generalize beyond training data and adapt to novel and unseen environments remains a critical challenge. It is further exacerbated for autonomous racing in which agents must be able to accurately perceive the environment, accounting for both the track layout and behavior of other agents, and quickly take actions that minimize lap times while avoiding collisions. In addition, modeling errors from uncertainties in the vehicle dynamics and noisy sensor measurements make it difficult to apply conventional planning and control algorithms. 

In recent years, learning-based control approaches that combine data-driven techniques with control theory have been successfully adopted for autonomous driving and related navigation tasks \cite{ugv_survey}. Model-based reinforcement learning (MBRL) is a subset of learning-based model predictive control (MPC) where the action policy is learned using reward signals provided from the agent's interaction with the environment. Unlike MPC which requires an a priori model of the system dynamics, the MBRL agent explicitly learns the world model through interactions. Learning the state transition probability from observed data enables the agent to adapt to changes in the environment or system dynamics, and account for uncertainties posed by imperfect knowledge of the operating conditions. The learned world model is then used to simulate future trajectories and plan actions that maximize the expected cumulative reward. 

\begin{figure}[t!]
  \centering
  \includegraphics[scale=0.38]{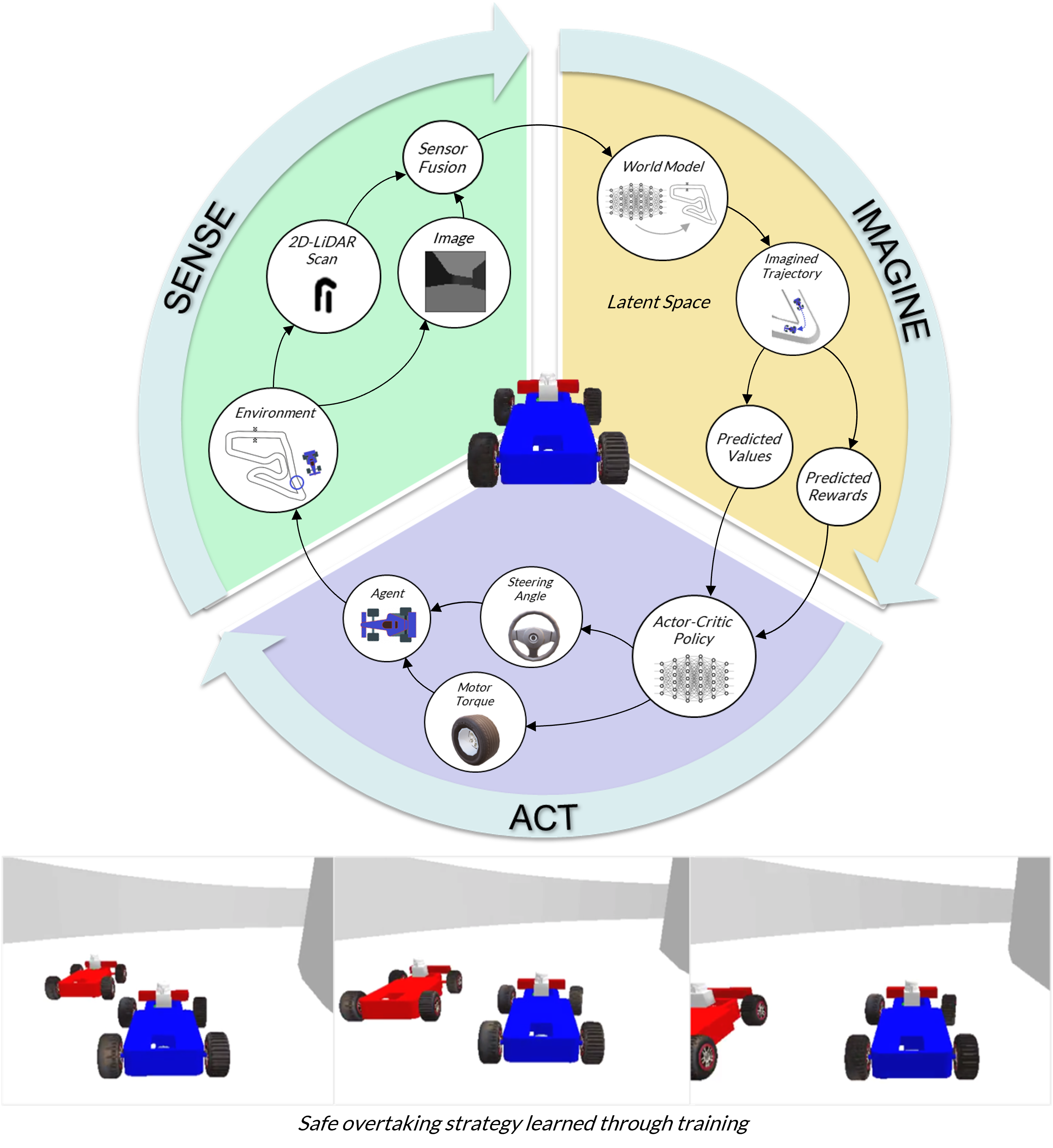}
  \caption{Overview of the proposed Sense-Imagine-Act paradigm for training multimodal model-based reinforcement learning agents for autonomous racing. Agents were trained in the F1TENTH Gym \cite{racingdreamer} against static obstacles and evaluated in a zero-shot head-to-head race against a fine-tuned rule-based agent. The resulting multimodal Dreamer agent learned a safe overtaking strategy while optimizing for speed and safety. }
  \label{overview}
  \vspace{-1em}
\end{figure}

Compared to model-free reinforcement learning (RL) approaches which learn a policy by directly mapping observations to actions, MBRL methods first capture a reduced-order representation of the environment or system dynamics from high-dimensional observations (e.g., images). The world model is abstracted to a latent state space, which is a multi-dimensional space that encodes salient features of the environment relevant for decision-making. The policy is then learned end-to-end by mapping between the latent space representation and actions (Fig.~\ref{overview}). As a result, MBRL can be more sample-efficient. Importantly, because planning with the learned world model can illuminate factors influencing the agent's behavior, MBRL is more interpretable than model-free RL when using high-dimensional observations \cite{mbrl_survey}. 

One of the key challenges in deploying MBRL agents is that the performance of the learned policy depends on the accuracy of the learned world model. To improve state estimation for autonomous racing, this paper explores methods for learning-based sensor fusion of egocentric 2D-LiDAR and RGB-camera observations. Multimodal perception is achieved by learning a joint representation of the sensor measurements in the latent state space.  To evaluate the accuracy of the world model learned using various modalities, this paper qualitatively and quantitatively evaluates the robustness of the learned policy on unseen race tracks and against a rule-based agent in a head-to-head race.   

The contributions of this work are the following:
\begin{enumerate}
    \item To the best of the authors' knowledge, this is the first end-to-end implementation of a multimodal MBRL agent for autonomous racing. Multimodal perception is learned via self-supervised sensor fusion of high-dimensional LiDAR and camera observations collected from the F1TENTH Gym (Fig.~\ref{overview}).
    \item Multimodal Dreamer agent that learns a hierarchical representation of the world model by fusing individual latent distributions of sensing modalities into an intermediate joint distribution using stacked encoders.
    \item Zero-shot head-to-head racing benchmark of sensing modalities (LiDAR, camera, and multimodal) against a rule-based agent representing a dynamic obstacle. Agents are trained against static obstacles and evaluated in a head-to-head race on the same track. 
\end{enumerate}

We plan to release the open-source code and model parameters for reproducibility, and provide benchmark data. The following section summarizes related work in learning-based control for autonomous racing while Section~\ref{section:problem} provides a formal definition of the problem. Sections~\ref{section:worldmodel} and~\ref{section:behaviorlearning} provide technical details of the proposed method while Section~\ref{section:results} presents results of the single-agent and multi-agent experiments.    


\section{Related Work} \label{section:relatedwork}

Learning-based control for autonomous racing and related navigation tasks for unmanned ground vehicles (UGVs) has been an active area of research in the past few decades. We will provide a brief discussion of relevant RL and deep learning approaches  in this section while referring readers to \cite{ugv_survey} for a comprehensive survey.

\textbf{Model-based Reinforcement Learning (MBRL).} One approach to addressing autonomous racing challenges of modeling uncertainty in the environment and system dynamics, partial observability, and temporal abstraction for sequential decision-making is to combine learning and planning techniques \cite{mbrl_survey}. Hafner \textit{et al.} \cite{planet} introduced a recurrent state-space model (RSSM) for learning a world model from high-dimensional images and combined it with online planning using MPC for learning long-horizon behaviors. Dreamer \cite{dreamer, dreamerv2, daydreamer} and TD-MPC \cite{tdmpc} are MBRL algorithms that learn the world model using images but plan by latent imagination using an actor-critic algorithm for end-to-end learning. 

\textit{End-to-end Learning.} Brunnbauer \textit{et. al} \cite{racingdreamer} demonstrated sim2real transfer of Dreamer for single-agent autonomous racing using 2D-LiDAR rays instead of images for building the world model, and pre-training using expert demonstration from a state-of-the-art obstacle avoidance algorithm. Dwivedi \textit{et. al} \cite{iccas} extended \cite{racingdreamer} with a plan-assisted architecture that leverages planning in trajectory-space to improve exploration in single-agent races. We improve the accuracy of the world model in \cite{racingdreamer} and \cite{iccas} by augmenting Dreamer with multimodal perception using 2D-LiDAR and RGB-camera. We also demonstrate zero-shot transfer for both single-agent and multi-agent autonomous racing without requiring expert demonstration or an external path planner. Furthermore, \cite{badgr} and \cite{meger} demonstrated real-world application of vision-based MBRL for off-road driving in static environments.  

\textit{Multimodal Perception.} Tremblay \textit{et. al} \cite{tremblay} and Triest \textit{et. al} \cite{tartandrive} improved the world model component of MBRL with multimodal perception for off-road driving but did not demonstrate end-to-end learning. \cite{tremblay} extended RSSM \cite{planet} with multimodal observations and introduced a training scheme to handle missing modalities. \cite{tartandrive} used a gated recurrent unit (GRU) \cite{gru} to model the latent dynamics and benchmarked multimodal neural network architectures for predicting vehicle trajectories. We investigate multimodal sensor fusion techniques for predicting how the environment changes for autonomous racing, explore hierarchical representations, and demonstrate end-to-end learning.  

\textbf{Model-free Reinforcement Learning (RL).} Another approach is to learn a mapping between observations and actions by augmenting RL agents with additional policy learning methods \cite{residual, sonyai, eth}. While computationally efficient, these methods require fine-tuning of hyperparameters and are less generalizable to data outside the training distribution.  

\textbf{Imitation Learning (IL).} Instead of collecting data through interactions, agents can also learn through expert demonstrations \cite{pan,rus,imitation}. However, IL may converge to sub-optimal policies if agents are not trained with high-quality and diverse sets of annotated data from multiple sensors, which are typically difficult to acquire for autonomous racing. 


\begin{figure*}[t]
  \centering
  \includegraphics[scale=0.5]{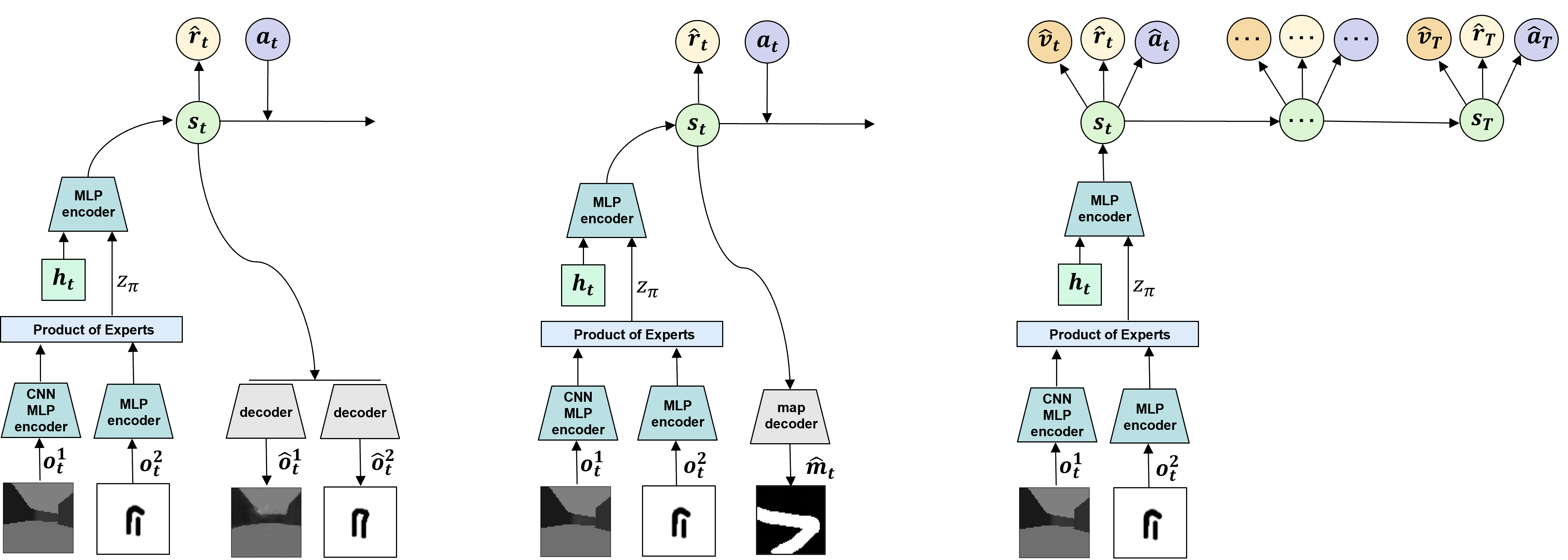}
  \caption{(\textit{left}) World model (\textit{s}) is learned using observations (\textit{o,\^o}) processed through a multimodal variational autoencoder and encoded actions (\textit{a}). (\textit{middle}) Supervised learning of the world model where the decoder constructs an occupancy grid (\textit{\^m}) instead of reconstructing the observations. (\textit{right}) The behavior learning component in the combined latent space optimizes policy using imagined trajectories with predicted state values (\textit{\^v}), rewards (\textit{\^r}), and actions (\textit{\^a}).}
  \label{dreamer}
  \vspace{-1em}
\end{figure*}

\section{Problem Definition} \label{section:problem}

Sequential decision-making problems are classically formalized as Markov decision processes (MDP) defined by a tuple $\langle  \mathcal{S}, \mathcal{A}, \mathcal{T}, \mathcal{R}\rangle$ containing the sets of states ($s_{t} \in \mathcal{S}$) and actions ($a_{t} \in \mathcal{A}$), the state transition function $\mathcal{T}(s_{t+1} | s_{t}, a_{t})$ which characterizes the environment's dynamics, and the reward function $\mathcal{R}(s_{t},a_{t},s_{t+1})$\cite{RLtext}. Given a trajectory $\tau = (s_{0},a_{0},r_{0}), \ldots, (s_{T},a_{T},r_{T})$, the goal of the RL agent is to maximize the discounted cumulative reward $\mathbb{E}_{\tau}(\sum_{t=0}^{T} \gamma^{t} r_{t})$, where $\gamma$ is the discount factor and $r_{t}$ is the learning signal from $\mathcal{R}$. The Markov property implies that the state encapsulates information about all previous interactions with the environment. However, the vehicle's state in robotics is typically estimated using imperfect observations collected from noisy sensors. In addition, the environment's dynamics in real-world scenarios are partially observable and non-stationary (i.e., stochastic).

To capture the uncertainty in the state estimation, the RL problem for autonomous racing is formulated as a Partially Observable MDP (POMDP) defined by a tuple $\langle  \mathcal{S}, \mathcal{A}, \mathcal{T}, \mathcal{R}, \Omega, \mathcal{O}\rangle$ \cite{pomdp}. The POMDP extends the MDP by containing the set of observations ($o_{t} \in \Omega$) and the observation function $\mathcal{O}(s_{t+1},a_{t},o_{t})$, which characterizes the probability of seeing an observation after taking an action that transitioned the environment into the new state. 

Autonomous racing in the F1TENTH Gym is a continuous control problem with normalized motor torque $\delta^{\omega}_{t} \in [0.005, 1]$ and normalized steering angle $\delta^{s}_{t} \in [-1, 1]$ as actions \cite{racingdreamer}. Physically, the vehicle is able to achieve a maximum velocity of 5m/s and steering angle of $\pm24^\circ$. Observations are raw sensor measurements collected in first-person view and rewards are pre-defined and deterministically based on the state-action pair. Given a trajectory, the MBRL agent must first learn the transition function that computes the belief state or the distribution over the latent states that capture the history of the environment's dynamics. Afterwards, the agent must learn a policy that enables it to safely traverse the race track while minimizing the lap time. The following section describes how the belief state is incrementally updated using variational inference. 


\section{World Model} \label{section:worldmodel}

The world model is a combined latent state space consisting of compact representations of high-dimensional observations collected from multiple sensing modalities. Multimodal Dreamer relies on latent vectors encoded using a multilayer perceptron (MLP) encoder for the 2D LiDAR rays (1080 x 1) covering a $270^{\circ}$ field-of-view (FOV), and a convolutional neural network (CNN) for the low resolution RGB camera images (64 x 64).  

\subsection{Multimodal Perception}
The recurrent state-space model (RSSM) is a time-series latent variable model with deterministic and stochastic states \cite{planet,dreamer}. In this work, RSSM is extended to include a set of observations $O_{t} = (o_{t}^{M},\ldots,o_{T}^{M})$ where $M$ represents the sensing modality (egocentric LiDAR or camera observations). RSSM learns the latent dynamics of the system through the representation, transition, observation, and reward models. These models are Gaussian with mean and variance parameterized by deep neural networks jointly updated by the parameter $\theta$:
\begin{align}
\begin{split}\label{eq:1}
    Representation:{}& \hspace*{1mm} p_{\theta}(s_{t}|s_{t-1},a_{t-1},O_{t})   \backsim\mathcal{N}(\mu,\Sigma),
\end{split}\\
\begin{split}\label{eq:2}
    Transition:{}& \hspace*{1mm} q_{\theta}(s_{t}|s_{t-1},a_{t-1})   \backsim\mathcal{N}(\mu,\sigma^2),
\end{split}\\
\begin{split}\label{eq:2}
    Observation:{}&  \hspace*{1mm} q_{\theta}(O_{t}|s_{t})  \backsim\mathcal{N}(\mu,\mathbb{I}),
\end{split}\\
    Reward:{}& \hspace*{1mm} q_{\theta}(r_{t}|s_{t})  \backsim\mathcal{N}(\mu,1).
\end{align}

\textbf{Self-supervised Learning.} As previously described, the 2D LiDAR rays are encoded and decoded through MLP networks while the camera images are encoded through a CNN and decoded through a transposed CNN. Due to the difference in sensing modalities, observations are not processed through a single variational autoencoder as implemented in \cite{meger} with top-down and forward-facing images or simply concatenated as done in \cite{daydreamer}. Instead, the observation model is implemented using a multimodal variational autoencoder (MVAE) where each modality is assumed to be conditionally independent \cite{mvae}. As a result, the representation model uses RSSM combined with MVAE, the transition model also uses RSSM, and the reward model uses a dense network.  

The belief state ($s_{t}$) is approximated using the product-of-experts (PoE) formulation \cite{poe} with the representation model as the joint posterior and the transition model as the prior expert. Multimodal Dreamer uses the PoE formulation to combine the observations into an intermediate latent space ($z_{\pi}$). Because all of the distributions are Gaussian, the PoE of the observations are analytically computed using the means and standard deviations of the individual modalities \cite{tartandrive}:
\begin{equation}
\label{eq:5}
\prod_{o_{t}^{M} \in O_{t}} q(s_{t}|o_{t}^{M}) = \mathcal{N}\left(\dfrac{\sum_{M} \left(\dfrac{\mu_{M}}{\sigma_{M}}\right)}{\sum_{M} \left(\dfrac{1}{\sigma_{M}}\right)}, \mathbb{I}\left( \sum_{M} \dfrac{1}{\sigma_{M}}\right)\right).
\vspace{1mm}
\end{equation}

The output of the PoE is then fused with the deterministic component ($h_{t}$) of the prior distribution using an encoder that parameterizes the posterior distribution (Fig.~\ref{dreamer}). Finally, the low-dimensional belief state is sampled from the encoded distribution. The stacked encoder design and sensor fusion using an intermediate latent state space enables multimodal Dreamer to learn a hierarchical representation of the environment. Without the intermediate latent space, the PoE formulation gives significant weight to high-dimensional observations, emphasizing images over LiDAR rays. \cite{muse} previously demonstrated that hierarchical representations improve the robustness of state estimation using multimodal observations and outperform variational autoencoders in single and cross-modality reconstructions. Section~\ref{section:singleagent} presents the evaluation results of two variations of multimodal perception: (1) PoE-based stacked encoder that learns a hierarchical representation (multimodal Dreamer) and (2) a single encoder with concatenated observations (multi-RSSM Dreamer). 

\textbf{Supervised Learning.} Supervised learning provides an alternative approach to learning the world model. Instead of reconstructing the original observations, the decoder predicts a local occupancy grid based on the agent's belief state which is encoded using both LiDAR and camera observations. The occupancy grid is constructed using a multivariate Bernoulli distribution that captures the probability of pixels being occupied, and trained using a bird's eye view of the local map \cite{racingdreamer}. The resulting agent is referred to as multimodal Dreamer (map) and benchmarked against other MBRL agents in the single-agent race.     

\subsection{Training Objective}

Models described in \eqref{eq:1}--\eqref{eq:5} are jointly optimized to increase the evidence lower bound (ELBO). ELBO for the self-supervised multimodal Dreamer includes reconstruction terms for each of the sensing modalities, a reward loss, and a Kullback-Leibler (KL) divergence regularizer for the approximate posterior extended for multimodal perception: 
\vspace{-1mm}
\begin{multline}
\text{ELBO}(O_{t}) = \mathbb{E}\bigg(\underbrace{\ln q_{\theta}(r_{t}|s_{t})}_{\text{Reward Prediction}} + \underbrace{\sum_{t} \beta_{M} \ln q_{\theta}(o_{t}^{M}|s_{t})}_{\text{Reconstruction}} + \\ 
\underbrace{-\beta_{KL} D_{KL}\big(p_{\theta}(s_{t}|s_{t-1},a_{t-1},O_{t}) || q_{\theta}(s_{t}|s_{t-1},a_{t-1})\big)}_{\text{KL Divergence Regularizer}}\bigg).
\end{multline}
\vspace{1mm}
ELBO for multimodal Dreamer (map) with supervised learning of the occupancy grid (ground truth $m_{t}$) replaces the observation reconstruction losses with an occupancy grid prediction loss, $\ln q_{\theta}(m_{t}|s_{t})$, conditioned on the belief state. 

The reconstruction losses are weighed by $\beta_{M}$, tuned to emphasize a particular modality, and the KL divergence regularizer $D_{KL}$ is weighed by $\beta_{KL}$. The regularizer measures the difference between the representation and transition models, providing a learning signal that minimizes the information gain of the observations on the latent dynamics. Overall, the model parameters $\theta$ are jointly optimized to maximize the likelihood of observations and rewards for a given state visited during training. 


\section{Behavior Learning} \label{section:behaviorlearning}

Given an encoded observation, the learned world model is used to generate imagined trajectories of states, rewards, and actions (Fig.~\ref{dreamer}). Multimodal Dreamer learns the policy in the combined latent space using an actor-critic network without decoding the observations. The actor network learns a policy, $\pi_{\phi}(a_{t}|s_{t})$, that aims to maximize the value estimates of the states while the critic network learns a value function, $v_{\psi}(s_{t})$, that aims to match the value estimates, providing a learning signal for the actor network. Both the actor and value models use a dense neural network parameterized by $\phi$ and $\psi$, respectively. The objective functions of the actor-critic networks include value estimates, $V_{\lambda}(s_{t})$, of the states in the imagined trajectory:
\begin{align}
\begin{split}
    actor:{}& \hspace*{1em} \max_{\phi} \mathbb{E}_{\theta,\phi} \bigg(\sum_{t=0}^{H} V_{\lambda(s_{t})}\bigg),
\end{split}\\
    critic:{}& \hspace*{1em} \min_{\psi} \mathbb{E}_{\theta,\phi} \bigg(\sum_{t=0}^{H} \dfrac{1}{2} \lVert v_{\psi}(s_{t}) - V_{\lambda(s_{t})}\rVert^{2}\bigg).
\end{align}
Value estimates are summed over an imagination horizon $H$, which is a pre-defined number of predictions over future time steps. A key advantage of Dreamer's actor-critic algorithm is that long-horizon behaviors are learned by backpropagating the value estimates through the latent dynamics, thereby achieving better gradient updates. Additional information on the learning objective and equations used for value estimation can be found in \cite{dreamer}.

\subsection{Single-Agent Reward Function}
The value estimate is a function of the reward function, which is pre-defined and deterministically calculated for a given state-action pair. In the single-agent scenario, the RL agent is trained using a dense reward function \cite{racingdreamer} that maximizes progress on the track \eqref{reward}. The normalized progress $p_{t} \in [0,1]$ for the F1TENTH Gym is calculated for each pixel on the track using a distance transform applied to the grid. The reward function also contains a collision penalty to deter the RL agent from colliding with the walls. 
\begin{equation}
 r_{t}^{single} =
\begin{cases}
100 * |p_{t} - p_{t-1}| & \text{Progress}, \\
- 1 & \text{Collision}. \\
\end{cases}
\label{reward}
\end{equation}

\subsection{Multi-Agent Reward Function} \label{section:multi-agent}

In the multi-agent scenario, the MBRL agent is trained against static obstacles (red cars) and evaluated against Waypoint Follower, a rule-based agent. Waypoint Follower operates at a fixed speed with steering angle guided by a PID controller that tracks pre-defined waypoints set equidistant between the track's wall. The rule-based agent takes a deterministic path and functions as a dynamic obstacle for the MBRL agent.
\begin{equation}
 r_{t}^{multi} =
\begin{cases}
100 * |p_{t} - p_{t-1}| & \text{Progress}, \\
- 1 & \text{Collision}, \\
-0.1 *| \delta_{t}^{s} - \delta_{t-1}^{s} | & \text{Smooth Action}.\\
\end{cases}
\label{c1}
\end{equation}


The single-agent reward function was updated with an action regularizer \eqref{c1} to smooth steering angles $\delta_{t}^{s}$ learned by the actor-critic policy \cite{mysore,bangbang}. Minimizing variations in steering angles intuitively leads to smoother trajectories and was empirically shown to improve lap times in multi-agent training. Overall, the reward function encourages agents to learn the latent dynamics and avoid collisions with both the track walls and five static obstacles that are randomly initialized on the grid. Position and orientation of these static obstacles are updated at each environment reset in order to provide a diverse set of observations for representation learning of the world model. 

\section{Simulation Setup \& Results} \label{section:results}

Experiments were conducted in both single-agent and multi-agent scenarios to address the following key questions:
\begin{enumerate}
    \item How do individual modalities and multimodal perception impact performance of Dreamer in single-agent and head-to-head autonomous racing? 
    \item Does hierarchical representation improve the accuracy of the world model? 
    \item Does supervised learning of the occupancy grid improve performance of multimodal Dreamer?
\end{enumerate}

\subsection{Training Pipeline}

Experiments were run using the F1TENTH Gym which leverages the PyBullet physics engine for simulation. 2D-LiDAR scans with range of 15m and a 270$^\circ$ FOV were sampled at 25Hz. Low resolution images from a RGB camera with a 90$^\circ$ FOV were sampled at 100Hz.    

\textbf{Prefill Strategy.} The prefill stage initializes the dataset with trajectories $\tau = (s_{0},a_{0},r_{0}), \ldots, (s_{T},a_{T},r_{T})$ for 5,000 timesteps. All training runs use a random prefill strategy whereas \cite{racingdreamer} used a rule-based Follow-the-Gap (FTG) agent \cite{ftg} to collect ``expert" data for pre-training. FTG is a gap-based obstacle avoidance algorithm that previously won the 2019 F1TENTH Autonomous Grand Prix. The following experiments evaluated the performance of sensing modalities on end-to-end MBRL agents without the need for pre-training so that insights could be extended to domains where expert demonstrations are difficult to acquire. 

\textbf{Setting.} Each training episode is terminated after 2000 simulation steps (20s) or automatically after a collision. Similarly, evaluation episodes are terminated after 4000 simulation steps (40s) or automatically after a collision. 

\textbf{Hyperparameters.} Multiple intermediate latent space dimensions, $dim(z_{\pi}) = 200, 514, 1024,$ and $1080$, were evaluated, with the latter yielding the best results. ELBO weights were set to $\beta_{M} = 1$ and $\beta_{KL} = 1$. Comprehensive tuning of hyperparameters is left to future work; only the action repeat ($AR = 4$) parameter was adjusted for the camera-based and multimodal agents from hyperparameters used in \cite{racingdreamer} for LiDAR agents ($AR = 8$) due to discrepancies in the simulation update rate from using computationally expensive camera observations. 


\begin{figure}[t]
  \centering
  \includegraphics[scale=0.55]{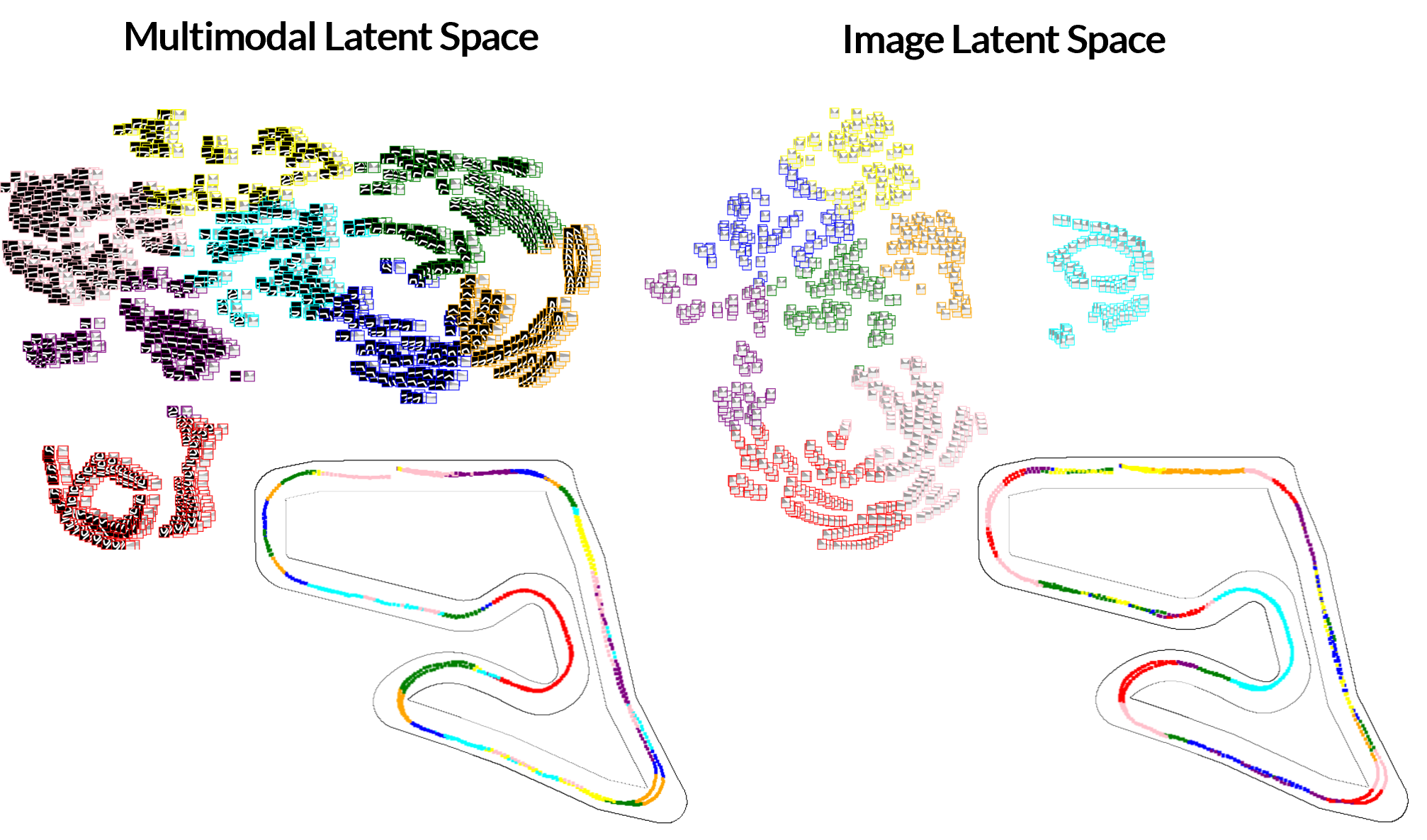}
  \caption{(\textit{left}) t-SNE of the combined latent space with camera and LiDAR observations (converted to occupancy grid for interpretability) overlapped and clustered with respect to the grid position on Austria. (\textit{right}) t-SNE of the camera's latent space and corresponding grid. Multimodal latent space captured features that led to a finer segmentation of turns (3 clusters).}
  \label{tsne}
\end{figure}

\begin{figure}[t]
\centering
\includegraphics[width=0.46\textwidth]{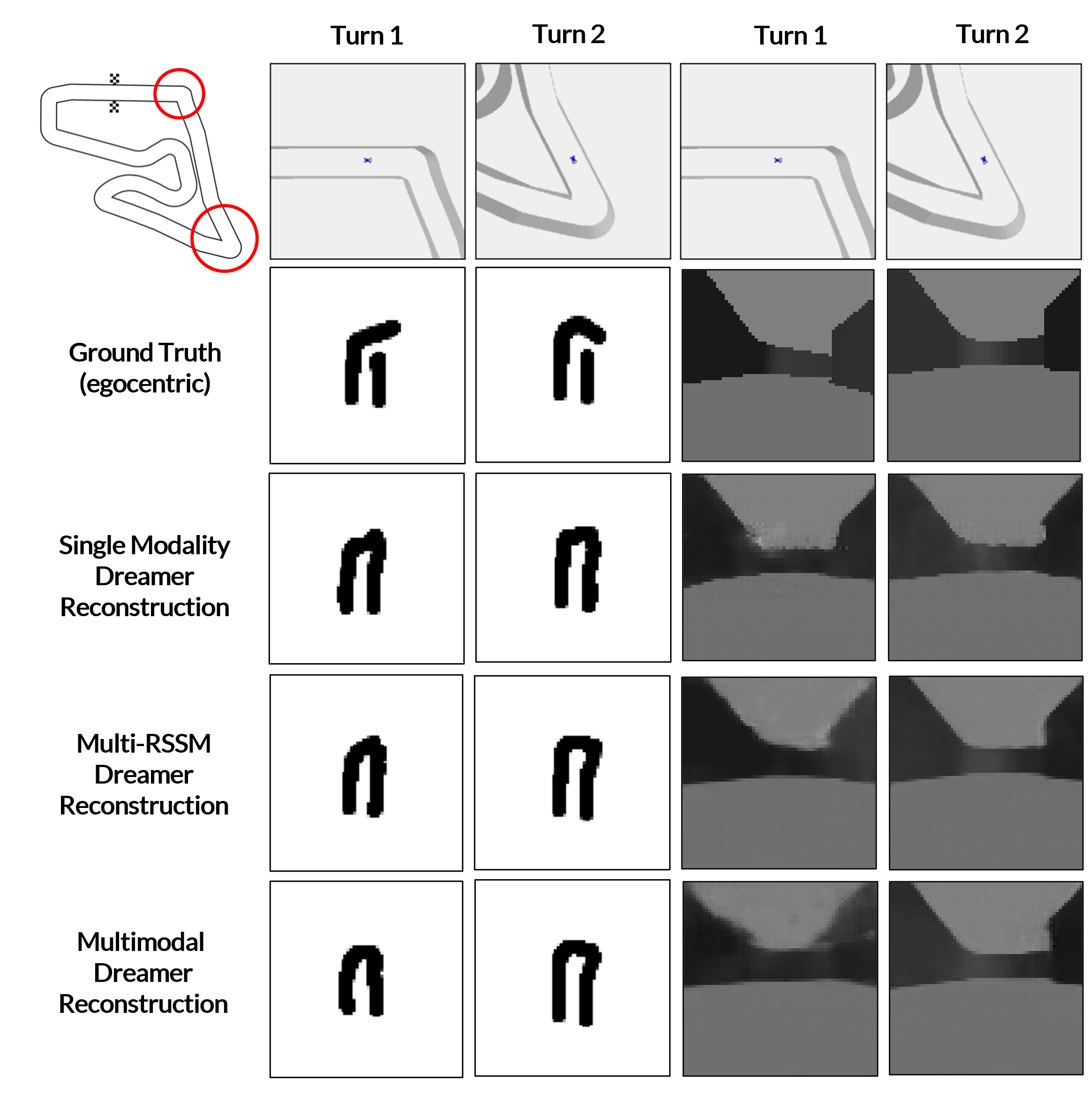}
  \caption{Qualitative comparison of single and multimodal reconstructions against ground truth collected at key points along Austria.  \label{qualeval} \vspace{-1em}}
\end{figure}

\begin{table} 
\vspace{1em}
  \centering
  \caption{Comparison of reconstruction quality using Structural Similarity Index (SSIM) and Cosine Similarity (COS) metrics. } \label{table:recon}
\begin{tabular}{ |p{1.1cm}|p{1.6cm}||p{0.7cm}|p{0.7cm}|p{0.7cm}|p{0.7cm}|  }
 \hline 
  \multicolumn{2}{|c||}{}  & \multicolumn{2}{c}{LiDAR} & \multicolumn{2}{c|}{Camera} \\
  \multicolumn{1}{|c}{Comparison} & \multicolumn{1}{c||}{Dreamer }& \multicolumn{1}{l}{SSIM} &  \multicolumn{1}{l}{COS} & \multicolumn{1}{l}{SSIM} &  \multicolumn{1}{l|}{COS} \\
 \hline
 \multirow{3}{*}{\parbox{1.2cm}{\centering Ground Truth vs. Turn 1}} &  Unimodal &  \textbf{0.9963} & \textbf{0.9993} &  \textbf{0.9614} & \textbf{0.9938}\\
 & Multi-RSSM  &  0.9959 & 0.9991 & 0.9513 & 0.9896\\
   &  Multimodal  & 0.9936  & 0.9985 & 0.9379 & 0.9862\\
 \hline
  \multirow{3}{*}{\parbox{1.2cm}{\centering Ground Truth vs. Turn 2}} &  Unimodal &  0.9875 & 0.9994 & 0.9875 & 0.9991\\
   & Multi-RSSM  &  \textbf{0.9975} & \textbf{0.9996} & \textbf{0.9909} & \textbf{0.9995}\\
   & Multimodal    & 0.9973  & 0.9995 & 0.9898 & 0.9993\\
  \hline
   \multirow{4}{*}{\parbox{1.2cm}{\centering Turn 1 vs. Turn 2}}  &  Unimodal &  0.9965 & \textbf{0.9994} &  0.9651 & 0.9959\\
 & Multi-RSSM   & 0.9959 & 0.9996 &  0.9592 & 0.9954\\
   &  Multimodal  & \textbf{0.9929} & 0.9995  & \textbf{0.9541} & \textbf{0.9938}\\
   \cline{2-6}
   &  \textit{Ground Truth} &   0.9956 & 0.9991 & 0.9392 & 0.9880\\
 \hline
\end{tabular}
\vspace{-1em}
\end{table}

\subsection{Single-Agent Race} \label{section:singleagent} 

All MBRL agents were trained on Austria for 2M timesteps and evaluated in Barcelona, Columbia, and Berlin tracks for 30 episodes. Pure model-free agents were considered but did not perform well on Austria \cite{racingdreamer, residual}, and therefore have been excluded from benchmarks. Agents were randomly initialized on the track during training in order to prevent overfitting to specific segments of the track. Distributed training was completed across multiple servers with NVIDIA Tesla V100 SXM2 and RTX A6000 GPUs, and Intel Xeon E5-2698 v4 and AMD Ryzen 9 5950X CPUs. Evaluation for all agents was performed on a laptop with a single GeForce GTX 1660 GPU and AMD Ryzen 5 3600 CPU.

\textbf{Qualitative Results.} Figure~\ref{tsne} shows the resulting latent space clustering for multimodal perception (left) and camera observations (right), colored with the corresponding position on the Austria race track. Observations were processed through a t-Distributed Stochastic Neighbor Embedding (t-SNE) \cite{tsnepaper}, a non-convex technique for converting high-dimensional Euclidean distances into conditional probabilities that represent similarities between neighboring data points. The resulting visualization is a two-dimensional map of high-dimensional inputs and preserves the number of data points before compression. 

LiDAR rays were converted to occupancy grids to distinguish between data points within the clusters, but encoded in the combined latent space using the raw scans. The size of clusters and distance between clusters may not provide any meaningful information due to limitations in interpreting t-SNE plots \cite{tsne}. However, the segmentation quality of the race track can be evaluated because it elucidates the relative difference in clustering between multimodal and camera latent spaces. It is evident from the color-coded Austria track that the multimodal latent space captures additional features from the shared representation, which enable it to further segment turns on the grid using three clusters (blue, orange, green) compared to two (red, pink) in the camera's latent space. The segmentation quality improves performance of the behavior policy that maps latent features to actions.  

Figure~\ref{qualeval} presents observations with the associated reconstructions by MBRL agents at various points along Austria while Table~\ref{table:recon} compares the reconstruction quality using Structural Similarity Index \cite{ssim} and Cosine Similarity \cite{cos} metrics. Multimodal agents are able to discern between the first (increasing radius) and second (sharp hairpin) turns in Austria while LiDAR Dreamer and camera Dreamer are likely to perceive them as similar segments. In addition, learning a hierarchical representation prevents the agent from overemphasizing high-dimensional observations, and instead allows it to focus on meaningful low-level features. For example, multimodal Dreamer infers the first turn to have a lower radius of curvature based on the relatively symmetric image reconstruction of the walls compared to the next turn.

\begin{figure}[t]
\includegraphics[width=.47\textwidth]{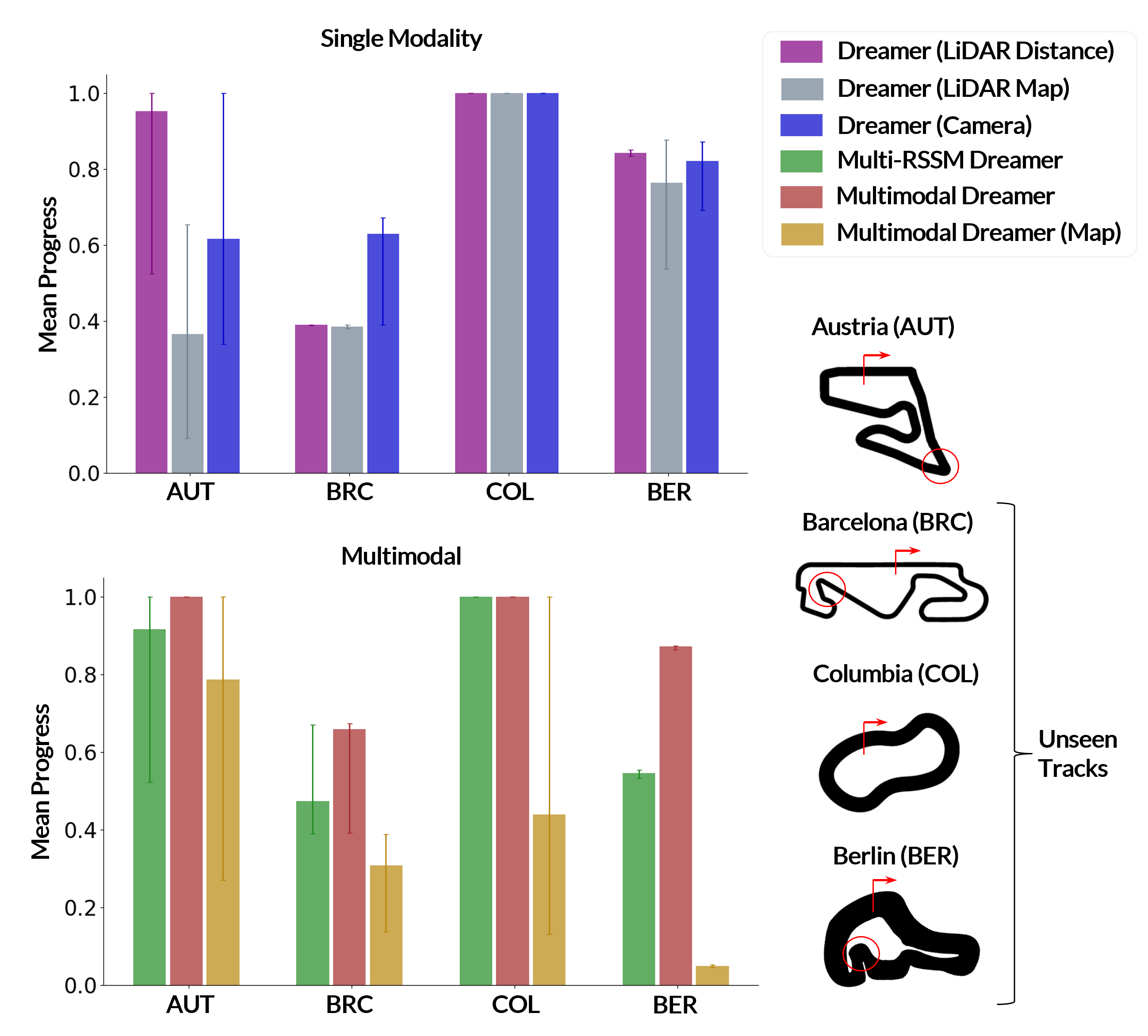}
  \caption{Evaluation results on Austria and zero-shot learning on unseen tracks. Bars denote mean progress over 30 episodes while delimiters show minimum and maximum progress. The prediction horizon was reduced from H = 15 to H = 5 for the unseen tracks. The best model for each agent was selected based on the mean progress from 5 seeded runs during training. \label{evalgraph}} 
  \vspace{-1em}
\end{figure}

\textbf{Quantitative Results.} While reconstructions from multi-RSSM Dreamer were closer to the ground truth, multimodal Dreamer achieved a higher mean progress on Austria (Fig.~\ref{evalgraph}). The ability to identify distinctive features of the track appears to be more impactful for improving performance of the policy than achieving high-quality reconstructions in the F1TENTH Gym. Though trained on Austria for 2M timesteps, all agents were able to generalize to Columbia. While none of the agents completed the remaining tracks, multimodal Dreamer achieved the best mean progress and low variance. Interestingly, self-supervised sensor fusion led to a higher mean progress compared to supervised learning of the occupancy grid. In addition, supervised learning solely with LiDAR scans generalized better than multimodal perception since it emphasized geometric features.  


\begin{figure*}[t]
  \centering
  \includegraphics[scale=0.5]{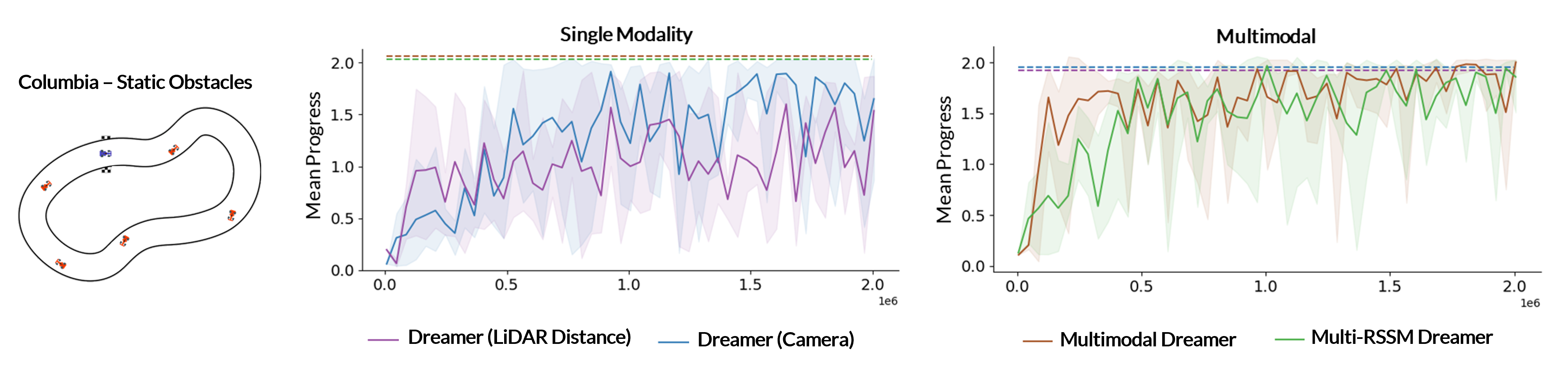}
    \vspace{-1em}
\caption{Multi-agent learning curves for single modality and multimodal agents on Columbia with static obstacles randomly initialized for each episode. Solid lines show the average mean progress from 5 seeded runs while dashed lines denote the highest mean progress achieved during training.}
  \label{multi-training}
\end{figure*}

\begin{table*} 
  \centering
  \caption{Multi-agent zero-shot evaluation (head-to-head racing on Columbia: 30 Trials, 1 Lap each)} \label{table:multi}
\begin{tabular}{ |p{3cm}||p{2cm}|p{2cm}|p{2cm}|p{2cm}|p{2cm}|p{2cm}|  }
 \hline 
  & Waypoint Follower &  LiDAR\hspace{3em} Dreamer \cite{racingdreamer} & Camera\hspace{3em}Dreamer \cite{dreamer} & Multi-RSSM\hspace{3em} Dreamer \cite{daydreamer} & Multimodal Dreamer  \\
 \hline
 Mean Progress (\%) & - & 60.22 & 83.40 & 99.35 & \textbf{100.0} \\
 \hline
  Mean / Best Lap Time (s) & 22.54 / 22.54 & 28.87 / 27.91 &  \textbf{21.20} / \textbf{20.53} & 21.37 / 20.91 & 21.64 / 20.85 \\
 \hline
  Mean Reward & - & 42.13 & 65.97  & 82.23 & \textbf{86.96} \\
  \hline
  Race Wins (\%) & - & 0.00 & {43.33} & 96.67 & \textbf{100.0}\\
 \hline
  Collision (\%) & - & 80.00 & 53.33 & 3.33 & \textbf{0.00} \\
 \hline
\end{tabular}
\end{table*}

\subsection{Multi-Agent Race}

All model-based RL agents were trained on Columbia for 2M timesteps against five randomly placed static obstacles (red cars) but evaluated against a rule-based agent described in Section~\ref{section:multi-agent} for 30 episodes. Distributed training for all agents was completed on a server with NVIDIA Tesla V100 SXM2 GPUs and Intel Xeon E5-2698 v4 CPUs. Evaluation was performed on a laptop with a single NVIDIA RTX 3060 GPU and Intel i7-12700H CPU. 

During evaluation, the rule-based agent is placed approximately 3m ahead of the MBRL agent on Columbia (minimum track width of 3.53m) in order to enable opportunities to overtake at various points along the track. Both agents are able to achieve top speeds up to 5m/s and have access to an identical range of control inputs. 

\textbf{Impact of Multimodal Perception.} Figure~\ref{multi-training} shows the learning curves (mean progress) for self-supervised single modality (LiDAR distance and camera) and multimodal agents on Columbia with five static obstacles randomly placed around the track. Both the LiDAR and camera-based agents converged to sub-optimal policies and on average completed less than two laps in 40s. The camera-based agent achieved a higher mean progress because it is able to discern between the static obstacles (red car) and the track walls. In comparison, both multimodal agents completed an average of two laps in 40s while multimodal dreamer achieved a faster convergence rate.  

Table~\ref{table:multi} summarizes the zero-shot evaluation results of the head-to-head race against a rule-based agent (Waypoint Follower). Out of 30 episodes, LiDAR dreamer had the highest collision rate and the slowest mean lap times of the completed races. The latent space representation of depth-based observations may not be able to discern between a dynamic obstacle and the track walls due to insufficient training data. For example, the LiDAR observation of a tight corner could potentially be similar to when the dynamic obstacle is positioned directly ahead of the MBRL agent. 

On the other hand, camera Dreamer achieved the best lap times on the track but remarkably only won 43\% of the races. Of the remaining races, 53\% prematurely ended due to collisions. The contrastive result highlights the delicate trade-off between maximising for speed and safety in head-to-head autonomous racing. While camera observations provide a higher spatial resolution of the environment, the FOV is limited to only 90$^\circ$ in front the agent. As a result, camera Dreamer is not as reactive as LiDAR Dreamer and will rigidly follow a racing line without adapting to the presence of the other agent on the track. 

Because multimodal perception leverages strengths of LiDAR and camera observations, both multimodal Dreamer and multi-RSSM Dreamer outperformed their single modality counterparts. The combination of multimodal perception and collision penalty in the reward function enabled the agent to optimize for speed and safety.      

\textbf{Impact of an Intermediate Latent Space.} Interestingly, multimodal Dreamer avoided collisions without employing an explicit safety layer over the policy. As discussed in Section~\ref{section:singleagent}, the intermediate latent space improves perception of meaningful low-level features.  Based on the mean lap time and collision rate, multi-RSSM Dreamer learned a more aggressive racing policy that maintains a tighter clearance from the walls and overtakes in close proximity to the dynamic obstacle. Furthermore, multimodal Dreamer also performed better in terms of the mean progress, best lap time, mean reward, and race wins.  

\subsection{Limitations}

While multimodal perception improved performance of Dreamer, it also incurred a higher computational cost (3x) of training with multiple high-dimensional observations. To improve real-time performance on a physical system, future work could investigate training schemes that encode latent distributions robust to missing modalities \cite{tremblay, muse} so that fewer computationally expensive camera observations would be required. Another limitation is the variability in the world model because the latent representations are only sampled once at each timestep. Future work could explore importance weighting strategies that exploit capabilities of variational inference in order to achieve a tighter ELBO \cite{vmloc}.

\section{Conclusion}

 Results showed that multimodal perception improves robustness of the world model and enables Dreamer \cite{racingdreamer,dreamer}, the state-of-the-art model-based reinforcement learning algorithm, to safely avoid collisions while minimizing lap times in zero-shot head-to-head autonomous racing. Our proposed method, multimodal Dreamer, learns a joint representation of 2D-LiDAR rays and high-dimensional images in the latent state space using the product-of-experts formulation \cite{mvae, poe}. Instead of using a single encoder \cite{daydreamer} to fuse all modalities, multimodal Dreamer first encodes observations into an intermediate latent space before further encoding the learned representation into a belief state.    
 
Although trained in Columbia with five static obstacles, multimodal Dreamer safely avoided collisions with a dynamic rule-based agent that was fine-tuned for the track. Zero-shot head-to-head racing performance suggests that the ability to identify meaningful low-level features is more impactful in improving performance of the policy than achieving high-quality reconstructions. In future work, we plan to augment the policy with a safety layer capable of correcting unsafe actions during online training in real-world environments.







\end{document}